\documentclass{article} 
\usepackage{nips12submit_e,times}
\usepackage{graphicx,amsmath,amsfonts,amssymb,bm, url, color, dsfont}
\usepackage{wrapfig}
\usepackage[small,bf]{caption}

\usepackage{hyperref}
\hypersetup{
    colorlinks=true,%
    citecolor=blue,%
    filecolor=blue,%
    linkcolor=blue,%
    urlcolor=blue
}

\title{Efficient Point-to-Subspace Query in $\ell^1$: Theory and Applications in Computer Vision\thanks{This work is published in the $12^{th}$ European Conference on Computer Vision (ECCV 2012)~\cite{sun2012l1}. A full version with technical details is available online \url{http://arxiv.org/abs/1208.0432}~\cite{sun2012l1_tr}.}}

\author{Ju Sun, Yuqian Zhang, and John Wright \\
Department of Electrical Engineering, Columbia University, NY, USA\\
\texttt{\{jusun, yuqianzhang, johnwright\}@ee.columbia.edu} \\
}

\nipsfinalcopy 

\begin{document}
\maketitle

\newtheorem{theorem}{Theorem}
\newtheorem{lemma}[theorem]{Lemma}
\newtheorem{corollary}[theorem]{Corollary}
\newtheorem{proposition}[theorem]{Proposition}
\newtheorem{definition}[theorem]{Definition}
\newtheorem{conjecture}[theorem]{Conjecture}
\newtheorem{problem}[theorem]{Problem}
\newtheorem{claim}[theorem]{Claim}
\newtheorem{remark}[subsection]{Remark}
\newtheorem{example}[subsection]{Example}

\newcommand{\eps}{\varepsilon}
\newcommand{\R}{\mathds{R}}
\newcommand{\Z}{\mathds{Z}}

\renewcommand{\Re}{\R}
\newcommand{\event}{\mc E}

\newcommand{\mb}{\mathbf}
\newcommand{\mc}{\mathcal}
\newcommand{\mf}{\mathfrak}
\newcommand{\md}{\mathds}
\newcommand{\mbb}{\mathbb}

\newcommand{\vtrz}{\mathrm{vec}}

\newcommand{\norm}[2]{\left\| #1 \right\|_{#2}}
\newcommand{\innerprod}[2]{\left\langle #1,  #2 \right\rangle}
\newcommand{\prob}[1]{\mbb P\left[ #1 \right]}
\newcommand{\expect}[1]{\mbb E\left[ #1 \right]}
\newcommand{\function}[2]{#1 \left(#2\right)}
\newcommand{\integral}[4]{\int_{#1}^{#2}\; #3\; #4}
\newcommand{\js}[1]{{\color{blue}{\bf Note: #1}}}
\newcommand{\jw}[1]{{\color{blue}{\bf John: #1}}}

\newcommand{\done}[2]{d_{\ell^1}\left(#1,#2\right)}
\newcommand{\bests}{\mc S_\star}

\begin{abstract}
Motivated by vision tasks such as robust face and object recognition, we consider the following general problem: \emph{given a collection of low-dimensional linear subspaces in a high-dimensional ambient (image) space and a query point (image), efficiently determine the nearest subspace to the query in $\ell^1$ distance}. We show in theory that Cauchy random embedding of the objects into significantly-lower-dimensional spaces helps preserve the identity of the nearest subspace with constant probability. This offers the possibility of efficiently selecting several candidates for accurate search. We sketch preliminary experiments on robust face and digit recognition to corroborate our theory. 
\end{abstract}

\section{Introduction}
Big data often come with prominent dimensionality and volume. Statistical learning and inference may be hard at such scales, due not only to conceivable computational burdens, but to stability issues. Nevertheless, parsimony that almost invariably dominates data-generating process dictates that structures associated with big data may be significant. For example, all images of a Lambertian convex object of fixed pose undergoing (remote) illumination changes lie approximately on a very low ($\approx 9$)-dimensional linear subspace, in the image space of dimensionality equal to the number of pixels per image (normally $\sim 10000$) \cite{Basri2003-PAMI}. Examples that have low-dimensional manifold structures abound (see, e.g., ~\cite{Roweis2000-Science, Donoho2005-JMIV}). By assuming reasonable structures for data, we can then turn the inference problem as an instance of nearest structure search, of which \emph{nearest subspace search} may serve as a basic building block: 
\begin{problem}[Nearest Subspace Search]
Given $n$ linear subspaces $\mc S_1, \dots, \mc S_n$ of dimension $r$ in $\Re^D$ and a query point $\mb q \in \Re^D$, determine the nearest $\mc S_i$ to $\mb q$. 
\end{problem}
Exploiting structures in big data has greatly helped in providing attractively simple formulations for learning and inference, and the remaining tasks are to make concrete the measure of ``nearness'' and to design efficient algorithm to solve the search problem.

\paragraph{Measure of Nearness.} Typically, one adopts a metric  $d(\cdot,\cdot)$ on $\Re^D$, and then sets $d(\mb q,\mc S_i) = \min_{\mb v \in \mc S_i} d(\mb q,\mb v).$ Certainly the appropriate choice of metric $d$ depends on our prior knowledge. For example, if the observation $\mb q$ is known to be perturbed by i.i.d.\ Gaussian noise from its originating subspace, minimizing the $\ell^2$ norm $d(\mb q,\mb v) = \| \mb q - \mb v \|_2$ yields a maximum likelihood estimator. However, in practice other norms may be more appropriate: particularly in situations where the data may have sparse but significant errors, the $\ell^1$ norm is a more robust alternative \cite{CandesE2005-IT, Wright2009-PAMI}. For images, such errors are due to factors such as occlusions, shadows, specularities. We focus on the choice of $\ell^1$ norm here and our main problem is
\begin{problem}[Main, Nearest Subspace Search in $\ell^1$]\label{prob:main}
Given $n$ linear subspaces $\mc S_1, \dots, \mc S_n$ of dimension $r$ in $\Re^D$ and a query point $\mb q \in \Re^D$, determine the nearest $\mc S_i$ to $\mb q$ in $\ell^1$ distance. 
\end{problem}

\paragraph{Efficiency of Algorithms.} We would like solve Problem~\ref{prob:main} using computational resources that depend as gracefully as possible on the ambient dimension $D$ and the number of models $n$, both of which could be very large for big data. The straightforward solution proceeds by solving a sequence of $n$ $\ell^1$ regression problems 
\begin{equation}  \label{eqn:L1}
d_{\ell^1}\left(\mb q, \mc S_i\right) = \min_{\mb v \in \mc S_i} \| \mb q - \mb v\|_1. 
\end{equation}
The total cost is $O( n \cdot T_{\ell^1}(D,r) )$, where $T_{\ell^1}(D,r)$ is the time required to solve~\eqref{eqn:L1}. The best known complexity guarantees for solving~\eqref{eqn:L1}, based on scalable first-order methods~\cite{Efron04leastangle,Beck2009-SJIS,Yin_bregmaniterative,Yang2010-ICIP}, are superlinear in $D$, though linear runtimes may be achievable when the residual $\mb q - \mb v_\star$ is very sparse \cite{Donoho06fastsolution} or the problem is otherwise well-structured \cite{AgarwalNIPS-2011}. So even in the best case, the straightforward solution has complexity $\Omega( n D )$. 
When both terms are large, this dependence is prohibitive: {\em Although Problem \ref{prob:main} is simple to state and easy to solve in polynomial time, achieving real-time performance or scaling massive databases appears to require a more careful study.}

\paragraph{Dealing with $D$ by Cauchy Random Embedding.} We present a very simple, practical approach to Problem \ref{prob:main} with much improved dependency on $D$\footnote{The reason that we do not deal with $n$ concurrently is discussed in Sec~\ref{sec:literature}. }. Rather than working directly in the high-dimensional space $\Re^D$, we randomly embed the query $\mb q$ and subspaces $\mc S_i$ into $\Re^d$, with $d \ll D$. The random embedding is given by a $d \times D$ matrix $\mb P$ whose entries are i.i.d.\ standard Cauchy's. That is to say, instead of solving \eqref{eqn:L1}, we solve
\begin{equation} \label{eqn:L1-proj}
d_{\ell^1}\left(\mb P\mb q, \mb P\mc S_i\right) = \min_{\mb v \in \mc S_i} \| \mb P \mb q - \mb P \mb v \|_1.
\end{equation}
We prove that if the embedded dimension $d$ is sufficiently large -- say $d = \mathrm{poly}( r \log n )$, then with constant probability the model $\mc S_i$ obtained from \eqref{eqn:L1-proj} is the same as the one obtained from the original optimization \eqref{eqn:L1}. The required dimension $d$ does not depend in any way on the ambient dimension $D$, and is often significantly smaller: e.g., $d = 25$ vs.\ $D = 32,000$ for one typical example of face recognition. The resulting (small) $\ell^1$ regression problems are amenable to customized interior point solvers (e.g., \cite{Mattingley}). The price paid for this improved complexity is a small increase in the probability of failure to locate the nearest subspace. Our theory quantifies how large $d$ needs to be to render this probability of error under control. Repeated trials with independent projections $\mb P$ can then be used to make the probability of failure as small as desired.  

\section{Cauchy Random Embedding and Theoretical Analysis}
Our entire algorithm relies on the standard Cauchy distribution with p.d.f. 
\begin{equation}
p_{\mc C}(x) = 1/\left[\pi \left(1+x^2\right)\right],  
\end{equation}
which is $1$-stable~\cite{uchaikin1999chance} and heavy-tailed (shown in Figure~\ref{fig:cauchy_dist}). The core of our algorithm is summarized as follows (right). \\
\\
\begin{minipage}{0.3\linewidth}
\includegraphics[width = 0.95\linewidth]{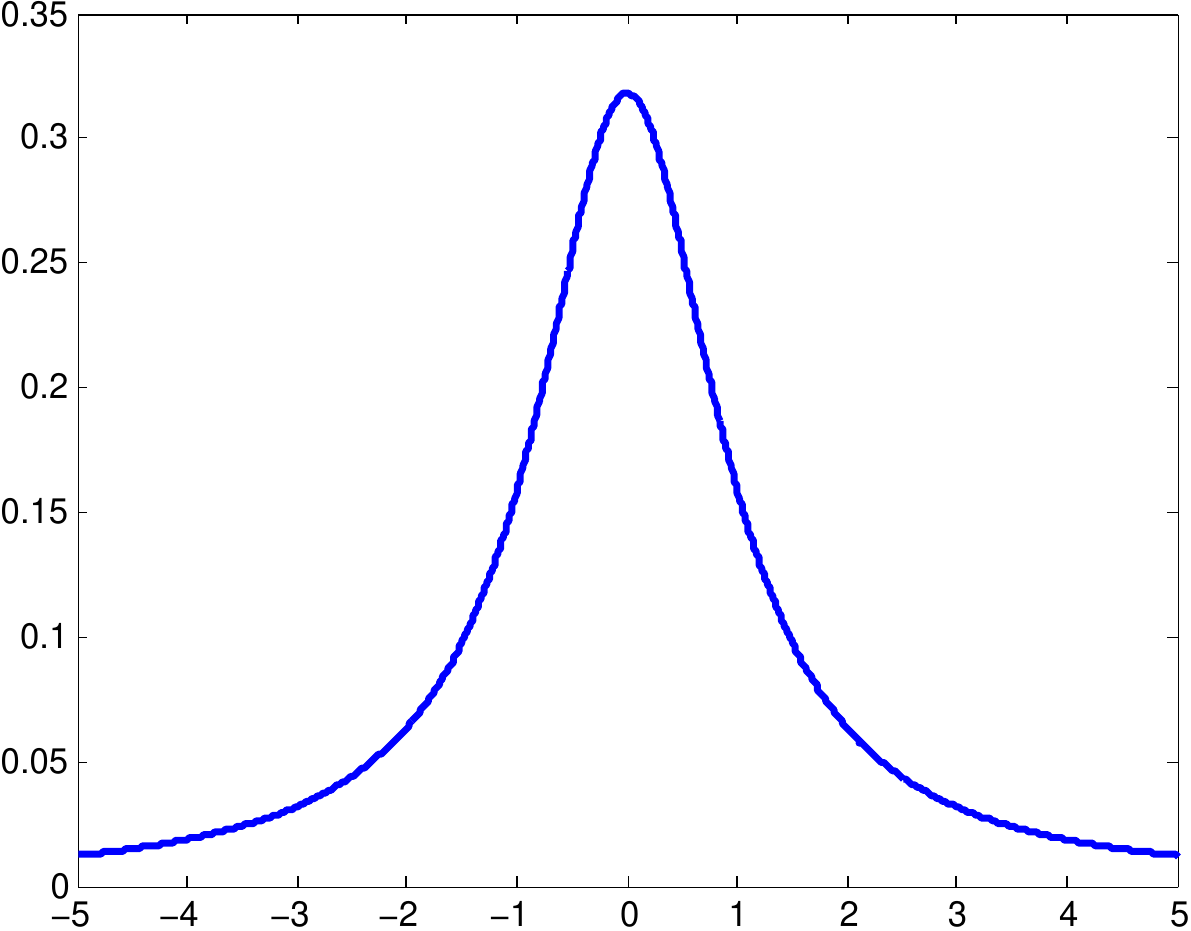}
\captionof{figure}{Standard Cauchy}
\label{fig:cauchy_dist}
\end{minipage} 
\hspace{0.05\linewidth}
\begin{minipage}{0.65\linewidth}
  \begin{tabular}{p{0.95\linewidth}}
      \hline
          \textbf{Input}: $n$ subspaces $\mc S_1, \cdots, \mc S_n$ of dimension $r$ and query $\mb q$ \\
          \textbf{Output}: Identity of the closest subspace $\mc S_\star$ to $\mb q$ in $\ell^1$ distance \\
          \hline \hline 
          \textbf{Preprocessing}: Generate $\mb P \in \R^{d\times D}$ with i.i.d.\ Cauchy RV's ($d \ll D$) and Compute the projections $\mb P\mc S_1$, $\cdots$, $\mb P \mc S_n$  \\
           \textbf{Test}: Compute the projection $\mb P \mb q$, and compute its $\ell^1$ distance to each of $\mb P \mc S_i$ \\
    \hline
     \end{tabular}
\end{minipage} 
\\
\\
Our main theoretical result states that if $d$ is chosen appropriately, with at least constant probability, the subspace $\mc S_{i_\star}$ selected will be the original closest subspace $\mc S_\star$:
\begin{theorem}\label{thm:main}
Suppose we are given $n$ linear subspaces $\left\{\mc S_1, \cdots, \mc S_n\right\}$ of dimension $r$ in $\R^D$ and any query point $\mb q$, and that the $\ell^1$ distances of $\mb q$ to each of $\left\{\mc S_1, \cdots, \mc S_n\right\}$ are $\xi_{1'} \leq \cdots \leq \xi_{n'}$ when arranged in ascending order, with $\xi_{2'}/\xi_{1'} \geq \eta > 1$. For any fixed $\alpha < 1 - 1/\eta$,  there exists $d \sim O\left[\left(r \log n\right)^{1/\alpha}\right]$ (assuming $n > r$), if $\mb P  \in \R^{d\times D}$ is iid Cauchy, we have
\begin{equation}
\mathop{\arg\min}_{i\in [n]} d_{\ell^1}\left(\mb P \mb{q}, \mb P \mc S_i\right) = \mathop{\arg\min}_{i\in [n]} d_{\ell^1}\left(\mb q, \mc S_i\right) 
\end{equation}
 with (nonzero) constant probability. 
\end{theorem}
At least two things are interesting about Theorem \ref{thm:main}: 1) $d$ depends on the relative gap $\eta$,  ratio of distances to the closest and to the second closest subspaces. Notice that $\eta \in [1,\infty)$, and that the exponent $1/\alpha$ becomes large as $\eta$ approaches one. This suggests that our dimensionality reduction will be most effective when the relative gap is nonnegligible; 2) $d$ depends on the number of models $n$ only through its logarithm. This rather weak dependence is a strong point, and, interestingly, mirrors the Johnson-Lindenstrauss lemma for dimensionality reduction in $\ell^2$, even though JL-syle embeddings are impossible for $\ell^1$. 

Additional practical implications of Theorem \ref{thm:main} are in order: 1) First, Theorem \ref{thm:main} only guarantees success with constant probability. This probability is easily amplified by taking a \emph{small number} of independent trials. Each of these trials generates one or more candidate subspaces $\mb S_i$. We can then perform $\ell^1$ regression in $\Re^D$ to determine which of these candidates is actually nearest to the query;  2) Since the gap $\eta$ is one important factor controlling the resource demanded, if we have reason to believe that $\eta$ will be especially small, we may instead set $d$ according to the gap between $\xi_{1'}$ and $\xi_{k'}$, for some $k' > 2$. With this choice, Theorem \ref{thm:main} implies that with constant probability the desired subspace is amongst the $k' - 1$ nearest (saved for further examination) to the query. If $k' \ll n$, this is still a significant saving over the naive approach. 

\begin{wrapfigure}[10]{r}{0.45\textwidth} 
  \vspace{-2pt}
    \includegraphics[width = 0.9\linewidth]{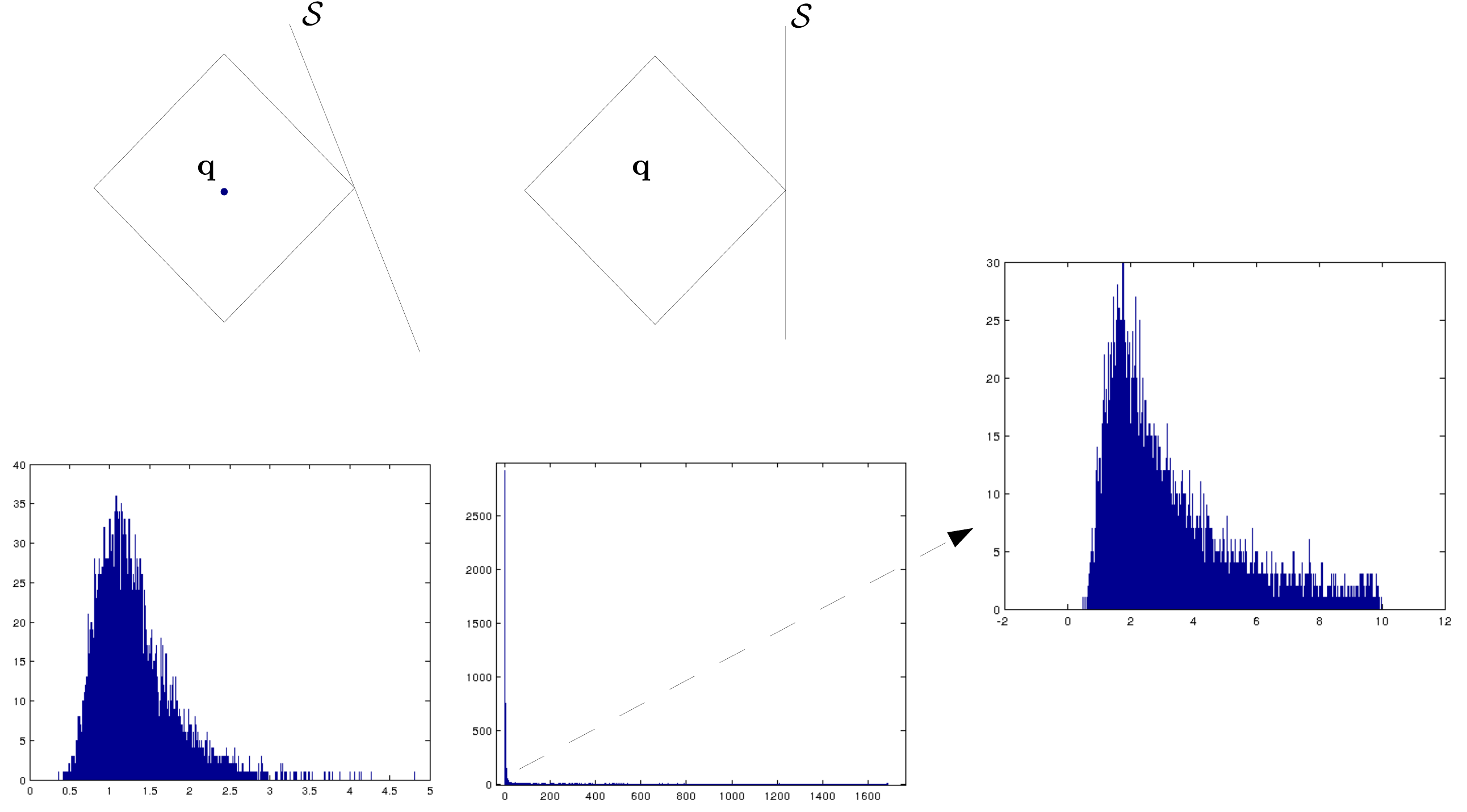}
    \captionof{figure}{An illustration of how random Cauchy embedding changes the query-to-subspace $\ell^1$ distance in statistics. }
    \label{fig:ratio_stat}
      \vspace{-20pt}
\end{wrapfigure}

\paragraph{Idea of the Analysis.} We present the full technical details in the report~\cite{sun2012l1_tr}, while highlight the intuition behind analysis now. Figure~\ref{fig:ratio_stat} shows a histogram of the random variable $\psi = \done{\mb P \mb q}{\mb P \mc S}$ ($\done{\mb q}{\mb S}$ is normalized), over randomly generated Cauchy matrices $\mb P$, for two different configurations of query $\mb q$ and subspace $\mc S$. 
Two properties are especially noteworthy. First, the upper tail of the distribution can be quite heavy: with non-negligible probability, $\psi$ may significantly exceed its median. In constrat, the lower tail is much better behaved: with very high probability, $\psi$ is not significantly smaller than its median. 
This inhomogeneous behavior (in particular, the heavy upper tail) precludes very tight distance-preserving embeddings using the Cauchy. However, our goal is {\em not} to find an (near-isometric) embedding of the data, per se, but rather to find the nearest subspace, to the query. In fact, it suffices to show that with nontrivial constant probability 
\begin{itemize} 
\item $\mb P$ does not increase the distance from $\mb q$ to $\bests$ too much;  and, 
\item $\mb P$ does not shrink the distance from $\mb q$ to any of the other subspaces $\mc S_i$ too much. 
\end{itemize} 
The observed inhomogeneous behavior is much less of an obstacle to establishing the desired results. 

\section{Related Work} \label{sec:literature}
Problem \ref{prob:main} is an example of a {\em subspace search} problem. In $\ell^2$, for $r=0$ and $r=1$, efficient algorithms with \emph{sublinear} query complexity in $n$ exist for the approximate versions~\cite{Datar04LSH, Andoni_approximateline}. For $r>1$, recent attempts~\cite{Basri2011-pami, Jain2010-NIPS} offered promising numerical examples, but not sublinear complexity guarantees. Results in theoretical computer science suggest that these limitations may be intrinsic to the problem~\cite{Williams05anew}. \\
\\
These attempts exploit special properties of the $\ell^2$ version of Problem \ref{prob:main}, and do not apply to its $\ell^1$ variant. However, the $\ell^1$ variant retains the aforementioned difficulties, suggesting that an algorithm for $\ell^1$ near subspace search with sublinear dependence on $n$ is unlikely as well.\footnote{Although it could be possible if we are willing to accept time and space complexity exponential in $r$ or $D$, ala \cite{Magen2008}.} This motivates us to focus on ameliorating the dependence on $D$. Our approach is very simple and very natural: Cauchy projections are chosen because the Cauchy is the unique $1$-stable distribution, a property which has been widely exploited in previous algorithmic work \cite{Datar04LSH,Li2007-JMLR,sohler2011subspace}. \\
\\
However, on a technical level, it is not obvious that Cauchy embedding should succeed for this problem. The Cauchy is a heavy tailed distribution, and because of this it does not yield embeddings that very tightly preserve distances between points, as in the Johnson-Lindenstrauss lemma. In fact, for $\ell^1$, there exist lower bounds showing that certain point sets in $\ell^1$ cannot be embedded in significantly lower-dimensional spaces without incurring non-negligble distortion \cite{Brinkman}. For a single subspace, embedding results exist -- most notably due to Soehler and Woodruff \cite{sohler2011subspace}, but the distortion incurred is so large as to render them inapplicable to Problem \ref{prob:main}.

\section{Experimental Verification}
Again we highlight part of our experiments here and more details can be found in the report~\cite{sun2012l1_tr}. We take the The Extended Yale B face dataset~\cite{Georghiades2001-PAMI} ($n=38$, $D = 168*192 \sim 30000$) and treat the facial images of one person as lying on a \emph{9-dimensional} linear subspace (as argued in~\cite{Basri2003-PAMI} and practiced in~\cite{Wright2009-PAMI}). For each subject, we take half of the images for training ($1205$ in total) and the others for testing ($1209$ in total). To better illustrate the behavior of our algorithm, we strategically divided the test set into two subsets: moderately illuminated ($909$, \textbf{Subset M}) and extremely illuminated ($300$, \textbf{Subset E}). \\
\\
Figure.~\ref{fig:EYB_moderate_vary_d} presents typical evolution of recognition rate on \textbf{Subset M} as the projection dimension ($d$) grows \emph{with only one repetition of the projection}. The high-dimensional NS (HDS) in $\ell^1$ achieves perfect ($100\%$) recognition, and the recognition rate (also probability of success as in Theorem~\ref{thm:main}) stays stable above $95\%$ with $d \geq 25$. Suppose the distance gap is significant such that $1/\alpha \to 1$, our theorem predicts $d = r \log n = 9 * \log 38 \approx 33$.  \\
\begin{minipage}{0.45\linewidth}
\begin{center}
\includegraphics[width = 0.9\linewidth]{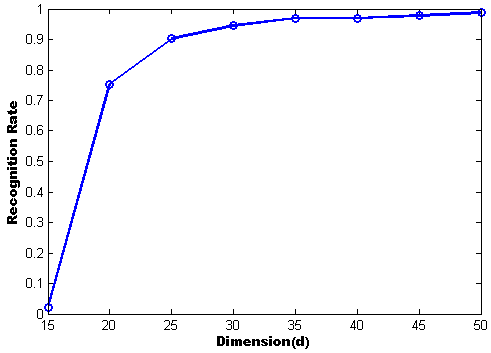}
\end{center}
\captionof{figure}{Recognition rate versus projection dimension ($d$) \emph{with one repetition} on \textbf{Subset M} face images of EYB.}
\label{fig:EYB_moderate_vary_d}
\end{minipage}
\hspace{1mm}
\begin{minipage}{0.5\linewidth}
\begin{center}
\includegraphics[width = 0.65\linewidth]{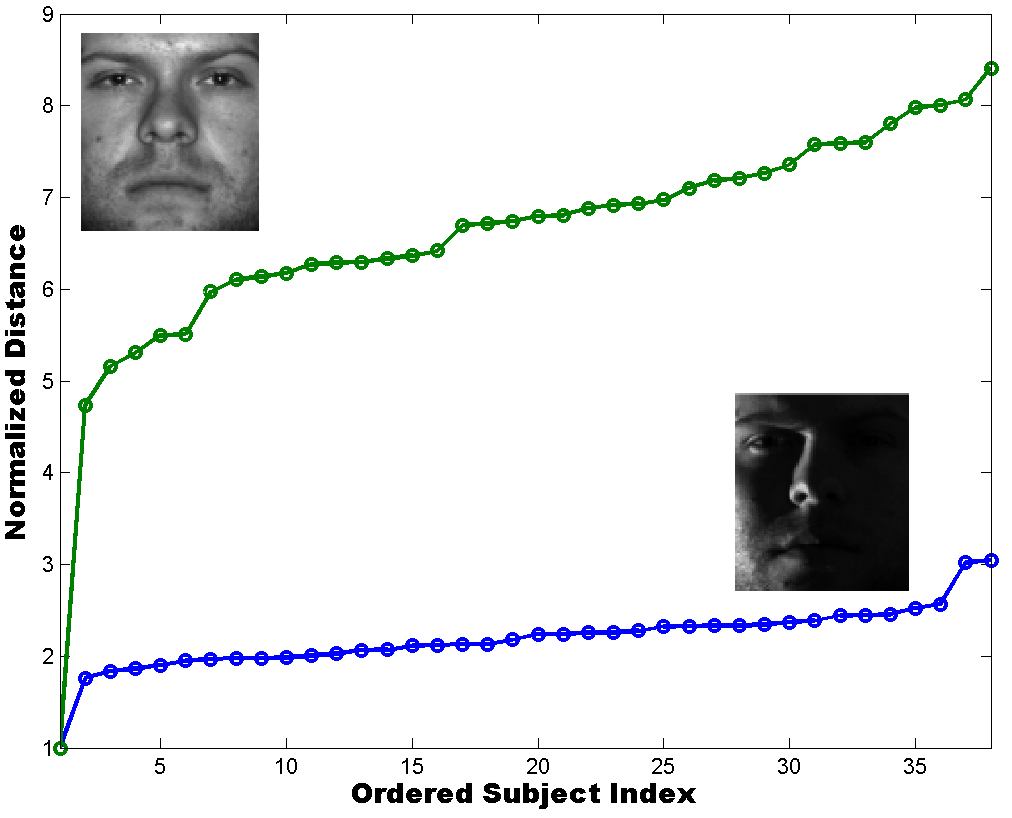}
\end{center}
\captionof{figure}{Samples of moderately/extremely illuminated face images and their $\ell^1$ distances to other subject subspaces.}
\label{fig:EYB_extreme_gap}
\end{minipage}\\
For extremely illuminated face images, the $\ell^1$ distance gap between the first and second nearest subspaces is much less significant (one example shown in Figure~\ref{fig:EYB_extreme_gap}). Our theory suggests $d$ should be increased to compensate for the weak gap (because the exponent $1/\alpha$ becomes significant). Our experimental results in Table~\ref{table:EYB_extreme_rate} confirm this prediction.   
\begin{table}[!htbp]
\caption{Recognition Rate on \textbf{Subset E} of EYB with varying $d$ and $N_{back}$ ($\#$ candidates for further test). }
\label{table:EYB_extreme_rate}
\centering
\begin{tabular}{c||lccc}
\hline
	  			&HDS	&$d=25$ 	&$d=50$		&$d=70$\\\hline
$r=15,N_{back}=5$	&94.7\%	&79.3\%	&87.7\%	&92.3\%\\\hline
$r=15,N_{back}=10$	&94.7\%	&87.3\%	&92.0\%	&94.0\%\\\hline
\end{tabular}
\end{table}

\newpage
{\small
\bibliographystyle{plain}
\bibliography{L1_Subspace_Search}

\begin{thebibliography}{10}

\bibitem{AgarwalNIPS-2011}
A.~Agarwal, S.~Negahban, and M.~Wainwright.
\newblock Fast global convergence of gradient methods for high-dimensional
  statistical recovery.
\newblock In {\em NIPS}, 2011.

\bibitem{Andoni_approximateline}
A.~Andoni, P.~Indyk, R.~Krauthgamer, and H.L. Nguyen.
\newblock Approximate line nearest neighbor in high dimensions.
\newblock In {\em SODA}, 2009.

\bibitem{Basri2011-pami}
R.~Basri, T.~Hassner, and L.~Zelnik-Manor.
\newblock Approximate nearest subspace search.
\newblock {\em {IEEE} Trans. PAMI}, 33(2):266--278, 2011.

\bibitem{Basri2003-PAMI}
R.~Basri and D.~Jacobs.
\newblock Lambertian reflectance and linear subspaces.
\newblock {\em {IEEE} Trans. PAMI}, 25(2):218--233, 2003.

\bibitem{Beck2009-SJIS}
A.~Beck and M.~Teboulle.
\newblock A fast iterative shrinkage-thresholding algorithm for linear inverse
  problems.
\newblock {\em SIAM J. on Imag. Sci.}, 2(1):183--202, 2009.

\bibitem{Brinkman}
Bo~Brinkman and Moses Charikar.
\newblock On the impossibility of dimension reduction in $\ell^1$.
\newblock {\em J. ACM}, 52:766--788, 2005.

\bibitem{CandesE2005-IT}
E.~Cand{{\' e}s} and T.~Tao.
\newblock Decoding by linear programming.
\newblock {\em {IEEE} Trans.\ {IT}}, 51(12):4203--4215, 2005.

\bibitem{Datar04LSH}
Mayur Datar and Piotr Indyk.
\newblock Locality-sensitive hashing scheme based on p-stable distributions.
\newblock In {\em SCG}, pages 253--262. ACM Press, 2004.

\bibitem{Donoho2005-JMIV}
D.~Donoho and C.~Grimes.
\newblock Image manifolds which are isometric to {E}uclidean space.
\newblock {\em J. of Math. Imag. and Vis.}, 23(1):5--24, 2005.

\bibitem{Donoho06fastsolution}
D.~Donoho and Y.~Tsaig.
\newblock Fast solution of $\ell^1$-norm minimization problems when the
  solution may be sparse.
\newblock {\em {IEEE} Trans.\ {IT}}, 54(11):4789--4812, 2008.

\bibitem{Efron04leastangle}
B.~Efron, T.~Hastie, I.~Johnstone, and R.~Tibshirani.
\newblock Least angle regression.
\newblock {\em Annals of Statistics}, 32:407--499, 2004.

\bibitem{Georghiades2001-PAMI}
A.~Georghiades, P.~Belhumeur, and D.~Kriegman.
\newblock From few to many: Illumination cone models for face recognition under
  variable lighting and pose.
\newblock {\em {IEEE} Trans. PAMI}, 23(6):643--660, 2001.

\bibitem{Jain2010-NIPS}
P.~Jain, S.~Vijayanarasimhan, and K.~Grauman.
\newblock Hashing hyperplane queries to near points with applications to
  large-scale active learning.
\newblock In {\em NIPS}, 2010.

\bibitem{Li2007-JMLR}
P.~Li, T.~Hastie, and K~Church.
\newblock Nonlinear estimators and tail bounds for dimension reduction in
  $\ell^1$ using cauchy random projections.
\newblock {\em JMLR}, 8:2497--2532, 2007.

\bibitem{Magen2008}
A.~Magen and A.~Zouzias.
\newblock Near optimal dimensionality reductions that preserve volumes.
\newblock In {\em APPROX-RANDOM}, pages 523--534, 2008.

\bibitem{Mattingley}
J.~Mattingley and S.~Boyd.
\newblock {CVXGEN}: A code generator for embedded convex optimization.
\newblock {\em Optimization and Engineering}, 13(1):1--27, 2012.

\bibitem{Roweis2000-Science}
S.~Roweis and L.~Saul.
\newblock Nonlinear dimensionality reduction by locally linear embedding.
\newblock {\em Science}, 290:2323--2326, 2000.

\bibitem{sohler2011subspace}
C.~Sohler and D.P. Woodruff.
\newblock Subspace embeddings for the $\ell_1$-norm with applications.
\newblock In {\em STOC}, 2011.

\bibitem{sun2012l1}
J.~Sun, Y.~Zhang, and J.~Wright.
\newblock Efficient point-to-subspace query in $\ell^1$ with application to
  robust face recognition.
\newblock In {\em ECCV}, 2012.

\bibitem{sun2012l1_tr}
J.~Sun, Y.~Zhang, and J.~Wright.
\newblock Efficient point-to-subspace query in $\ell^1$ with application to
  robust face recognition.
\newblock {\em CoRR}, abs/1208.0432, 2012.

\bibitem{uchaikin1999chance}
V.V. Uchaikin and V.M. Zolotarev.
\newblock {\em Chance and Stability: Stable Distributions and their
  applications}, volume~3.
\newblock Vsp, 1999.

\bibitem{Williams05anew}
Ryan Williams.
\newblock A new algorithm for optimal 2-constraint satisfaction and its
  implications.
\newblock {\em Theo. Comp. Sci.}, 348:357--365, 2005.

\bibitem{Wright2009-PAMI}
J.~Wright, A.Y. Yang, A.~Ganesh, S.S. Sastry, and Y.~Ma.
\newblock {Robust face recognition via sparse representation}.
\newblock {\em {IEEE} Trans. PAMI}, 31(2):210--227, 2009.

\bibitem{Yang2010-ICIP}
Allen Yang, Arvind Ganesh, Yi~Ma, and Shankar Sastry.
\newblock Fast $\ell^1$-minimization algorithms and an application in robust
  face recognition: A review.
\newblock In {\em ICIP}, 2010.

\bibitem{Yin_bregmaniterative}
W.~Yin, S.~Osher, D.~Goldfarb, and J.~Darbon.
\newblock Bregman iterative algorithms for $\ell^1$ minimization with
  applications to compressed sensing.
\newblock {\em SIAM J. Imag. Sci}, 1(1):143--168.

\end{thebibliography}
}

\end{document}